\documentclass[10pt,letterpaper,compsoc,conference]{iiswc24}

\usepackage{cite}
\usepackage{amsmath,amssymb,amsfonts}
\usepackage{algorithmic}
\usepackage{graphicx}
\usepackage[dvipsnames]{xcolor}
\usepackage[final]{microtype}
\usepackage[italic]{mathastext}
\usepackage{libertine}
\usepackage[T1]{fontenc}
\usepackage{textcomp}
\usepackage[varqu,varl]{zi4}
\usepackage[all]{nowidow}
\usepackage{authblk} % Required for multiple affiliations
\usepackage[keeplastbox]{flushend}
\usepackage{fancyhdr}

\usepackage{color}
\usepackage{microtype}
\usepackage{xspace}
\usepackage{graphicx}
\usepackage{subfigure}
\usepackage{booktabs} % for professional tables

\usepackage{multirow}
\usepackage{comment}

\usepackage{hyperref}
\usepackage{bm}

\usepackage{todonotes}

\usepackage{pifont}

\usepackage{amsmath,amsfonts}
\usepackage{algorithmic}
\usepackage{graphicx}
\usepackage{textcomp}
\usepackage{xcolor}
\usepackage{multirow}
\RequirePackage[nameinlink]{cleveref} % 'nameinlink' to mimic \autoref's behavior
\crefname{chapter}{Chapter}{Chapters}
\crefname{section}{Section}{Sections}
\crefname{subsection}{Section}{Sections}
\crefname{equation}{Equation}{Equations}
\crefname{definition}{Definition}{Definitions}
\crefname{assumption}{Assumption}{Assumptions}
\crefname{theorem}{Theorem}{Theorems}
\crefname{figure}{Figure}{Figures}
\crefname{table}{Table}{Tables}
\crefname{algorithm}{Algorithm}{Algorithms}
\let\autoref\cref % set \autoref as an alias for \cref

\newcommand{\insertFigure}[2]{
    \begin{figure}[t]
        \centering
        \includegraphics[width=\linewidth]{\FIGDIR/#1.pdf}
        \vspace{-1mm}
        \caption{#2}
        \label{fig:#1}
    \end{figure}
}

\newcommand{\insertWideFigureShrink}[3]{
    \begin{figure*}[h]
        \centering
        \includegraphics[width=#3\textwidth]{\FIGDIR/#1.pdf}
        \caption{ #2}
        \label{fig:#1}
    \end{figure*}
}

\newcommand{\insertFigureShrink}[3]{
    \begin{figure}[t]
        \centering
        \includegraphics[width=#3\linewidth]{\FIGDIR/#1.pdf}
        \caption{\small #2}
        \label{fig:#1}
    \end{figure}
}

\ifx\forsubmission\undefined
\newcommand{\TODO}[1]{\textcolor{red}{TODO: #1}}
\newcommand{\RP}[1]{\textcolor{green}{RP: #1}}
\newcommand{\HK}[1]{\textcolor{blue}{HK: #1}}

\newcommand{\fixme}[1]{{\color{red} {\textbf{#1}}}}

\else
\newcommand{\TODO}[1]{\textcolor{red}{}}
\newcommand{\RP}[1]{\textcolor{green}{}}
\newcommand{\HK}[1]{\textcolor{blue}{}}

\newcommand{\fixme}[1]{{\color{red} {}}} 

\fi

\newcommand{\squishlist}{
 \begin{list}{$\bullet$}
  { \setlength{\itemsep}{0pt}
     \setlength{\parsep}{3pt}
     \setlength{\topsep}{3pt}
     \setlength{\partopsep}{0pt}
     \setlength{\leftmargin}{1.5em}
     \setlength{\labelwidth}{1em}
     \setlength{\labelsep}{0.5em} } }

\newcommand{\squishlisttwo}{
 \begin{list}{$\bullet$}
  { \setlength{\itemsep}{0pt}
     \setlength{\parsep}{0pt}
    \setlength{\topsep}{0pt}
    \setlength{\partopsep}{0pt}
    \setlength{\leftmargin}{2em}
    \setlength{\labelwidth}{1.5em}
    \setlength{\labelsep}{0.5em} } }

\newcommand{\squishend}{
  \end{list}  }

\newcommand{\betterparagraph}[1]{\noindent \textbf{#1. }}

\newcommand{\rtgen}{\textsc{\textit{RTGen}}\xspace}

\newcommand{\xmark}{\ding{55}}%
\usepackage{placeins}
\def\FIGDIR{./Figures}
\begin{document}

%% EDIT TITLE BELOW

\title{Exploring the Dynamic Scheduling Space of Real-Time Generative AI Applications on Emerging Heterogeneous Systems}

% HK: Exploring dynamic scheduling space of real-time generative AI application on emerging heterogeneous systems

% RP : Exploring the dynamic scheduling of real-time concurrent GenAI inference workloads on heterogenous systems

%% DO NOT EDIT THE FOLLOWING

%\renewcommand\Authsep{\qquad}
%\renewcommand\Authand{\qquad}
%\renewcommand\Authands{\qquad}

%% EDIT AUTHOR LIST BELOW

% \author{Author1 Rachid Karami}
% \author{Chakshu Moar}
% \author{Sheng-Chun Kao}
% \author{Hyoukjun Kwon}

%\author{Author3 Name}
%\affiliation{Full Name of Awesome School}

%%% ALTERNATIVE FORMAT FOR MULTIPLE SCHOOLS:
%%% 
% \author{
% }
\author[1]{Rachid Karami}
\author[2]{Rajeev Patwari}
\author[1]{Hyoukjun Kwon}
\author[2]{Ashish Sirasao}

\affil[1]{University of California, Irvine}
\affil[2]{Advanced Micro Devices Inc.}

\maketitle
% \thispagestyle{firstpage}
% \pagestyle{plain}

%% EDIT YOUR PAPER'S CONTENTS BELOW
\begin{abstract}
The integration of generative AI models, particularly large language models (LLMs), into real-time multi-model AI applications such as video conferencing and gaming is giving rise to a new class of workloads: real-time generative AI (\rtgen).
These workloads combine the compute intensity and dynamic execution patterns of generative models with the stringent latency and concurrency constraints of real-time inference. 
To meet the diverse demands of \rtgen workloads, modern edge platforms increasingly adopt heterogeneous system-on-chip (SoC) architectures that integrate CPUs, GPUs, and NPUs.
Despite the potential of heterogeneous SoC, the scheduling space complexity and performance implications of \rtgen workloads on such platforms remain underexplored.

In this work, we perform a comprehensive characterization of \rtgen workloads on AMD’s latest heterogeneous SoC, Ryzen AI. 
We construct realistic multi-model scenarios inspired by industry use cases and profile model performance across all available backends.
Using this data, we evaluate five scheduling policies and their impact on both real-time metrics (e.g., deadline violation rate) and LLM performance (e.g., time-to-first-token and tokens-per-second). 
Our results show that scheduling decisions significantly affect workload performance (e.g., leading to a 41.7\% difference in deadline violation rates on average), and highlight the need for scheduling strategies that are aware of workload dynamics and hardware heterogeneity.
Our findings underscore the importance of workload-aware, dynamic heterogeneous scheduling in enabling high-performance, on-device \rtgen applications.
\end{abstract}

\begin{comment}
Advancements in generative large language models (LLMs) have enabled the integration of LLMs across a wide range of applications.
% 
Fueled by the increasing demands to deploy LLMs and generative models locally on edge devices, advances across the machine learning (ML) systems stack enable local deployment of these workloads. 
% 
Notably, the adoption of AI PCs based on heterogeneous architectures facilitates the high performance local deployment of generative AI.
%  
This trend unlocks the potential for an emerging class of applications \rtgen workloads, that integrate real-time (RT) and generative (Gen) workloads in the same application.

% 
Executing \rtgen workloads on heterogeneous CPU-GPU-NPU systems presents new challenges due to the diversity of model architectures, dynamic input patterns, and concurrency-induced resource contention.
% 
In this work, we define a suite of realistic \rtgen workload scenarios inspired by applications such as AI assistants, AI powered video conferencing, and media content creation. 
% 
We characterize these workloads across two state-of-the-art heterogeneous platforms and demonstrate how naive scheduling policies can result in suboptimal performance and deadline violations.
% 
Through detailed analysis, we highlight the importance of dynamic, workload- and hardware-aware scheduling for meeting quality of experience targets in edge deployment of generative AI. 
% 
Our findings provide insights into the evolving scheduling needs of hybrid real-time and generative ML systems.

\end{comment}
\section{Introduction}
\label{sec:intro}
% - Gen Ai success. 
% - LLM exceptional. 
\insertWideFigureShrink{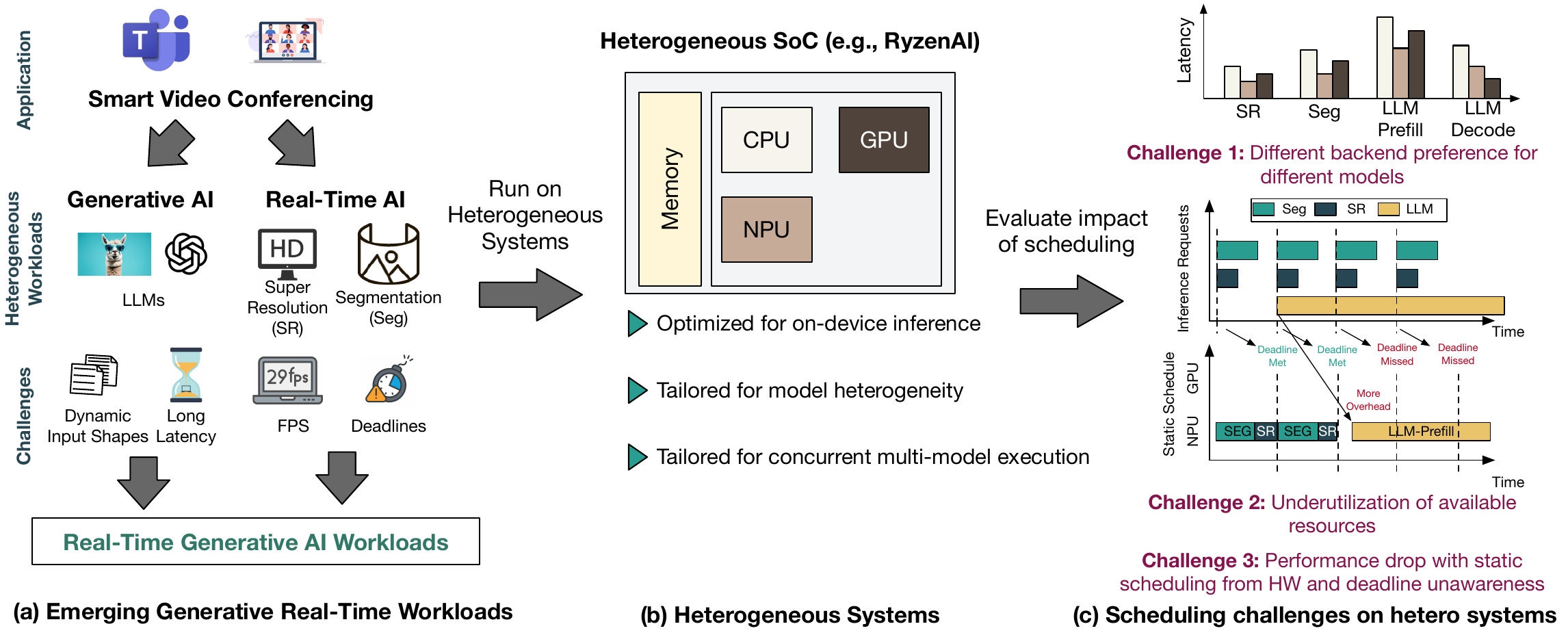}{Challenges in scheduling emerging real-time - generative workloads on heterogeneous systems.}{1}

The rapid advancement of artifical intelligence (AI) algorithm has successfully replaced many traditional algorithms with AI models such as those in computer vision~\cite{dosovitskiy2020image, liu2021swin} and language processing~\cite{vaswani2017attention}.
Such a trend led to recent real-time applications such as augmented and virtual reality (AR/VR) that combine multiple AI models for complex application functionality~\cite{kwon2023xrbench}.
One key trend today in such real-time multi-model AI workload is the inclusion of generative AI, which constructs real-time generative AI (\rtgen) workload.
For example,~\autoref{fig:overview-fig} (a) shows an example smart video conferencing application that utilizes a large language model (LLM) for real-time AI agent providing various functionalities such as summarization of the meeting and question answering based on the meeting context~\cite{zoomaiassistant}.
In addition to the large language model, other models such as image segmentation~\cite{zoom_blur_background} for background blurring are simultaneously executed in a real-time manner.
Such complex real-time AI workloads including generative AI models (i.e., \rtgen) lead to a major challenge to the AI system due to the heavy computation from generative AI and real-time requirement for other real-time AI models.
Furthermore, emerging \rtgen applications~\cite{kwon2023xrbench} demonstrate heavy demands on the on-device processing for better privacy, model customization opportunities, and low response time.

To provide desired on-device AI performance for the \rtgen, a new class of hardware platforms that integrate CPU, GPU, and NPU (Neural processing unit; AI accelerator) in one system-on-chip (SoC) has emerged~\cite{amd_ryzen_ai,apple_m3_chips, qualcomm_best_ai_pcs}, as illustrated in~\autoref{fig:overview-fig} (b).
Such heterogeneous SoCs, such as AMD Ryzen AI~\cite{amd_ryzen_ai}, provide not only extra parallelization opportunities across NPUs and GPUs for multi-model workloads~\cite{kwon2021herald, kwon2023xrbench} but also adaptivity to heterogeneous AI workloads in \rtgen.
For example, in our evaluation, Llama3-1B's prefill and decode stages are 64.7\% faster on NPU and 81.5\% faster on GPU, respectively.
%
%The data indicate that having NPU and GPU 

Although the heterogeneous SoC demonstrated promising features for \rtgen, one major challenge is exponentially growing scheduling space (e.g., for our video conference scenario on a CPU + GPU + NPU system, the scheduling space size is O({$2^{125}$})).
In addition, the scheduling decision has a significant impact on the resulting computational performance.
%
% For example, in our evaluation, we observe \TODO{XX\%} impact on \TODO{(metric name)}, on average.
%
However, the scheduling space and resulting performance horizon have not been well-studied, which hinders high-performance on-device \rtgen on heterogeneous SoCs.

Therefore, in this work, we conduct a thorough performance characterization study of \rtgen on one of the latest heterogeneous SoC, AMD Ryzen AI~\cite{amd_ryzen_ai}.
For that, we first construct a realistic \rtgen workload scenarios inspired by real use cases from industry.
Using the constructed workload scenarios, we conduct performance characterization of the usage scenarios on each backend on a heterogeneous SoC.
Finally, we apply various scheduling policies and delve into their impact on the real-time performance (deadline violation rate) and LLM performance (time to first token and the number of tokens per second).
Our characterization study shows that the impact of scheduling decision is significant (e.g., 41.7\% difference in  deadline violation rate, on average) and it is crucial to consider all the \rtgen's characteristics when we design a scheduler for \rtgen.
Also, our study shows that heterogeneous backends provide superior real-time and LLM performance compared to a homogeneous SoC, which motivates further investigation of the optimization opportunity in the intersection of \rtgen and heterogeneous SoCs.

We summarize our contributions as follows: 
\begin{itemize}
    \item {We define realistic \rtgen scenarios inspired by real-time applications integrating generative AI features, and highlight the challenges of this emerging class of multi-model workloads.}
    \item{We perform a workload characterization on a cutting edge heterogeneous platform, and show the importance of architectural heterogeneity for high performance deployment of \rtgen models on edge.}
    \item{We evaluate the impact of the scheduling policy on \rtgen workloads quantitatively, and highlight the scheduling challenges.}
    \item{We provide insights on the scheduling problem for \rtgen workloads running on heterogeneous systems.}
\end{itemize}

\section{Background and Motivation}
\label{sec:background}

\subsection{Large Language Models}
\label{subsubsec:llm}

% \insertWideFigure{LLM}{Overview of LLM architecture and generative inference.}
% \insertWideFigureShrink{overview-fig}{Challenges in scheduling emerging real-time - generative workloads on heterogeneous systems.}{1}

% 

\insertWideFigureShrink{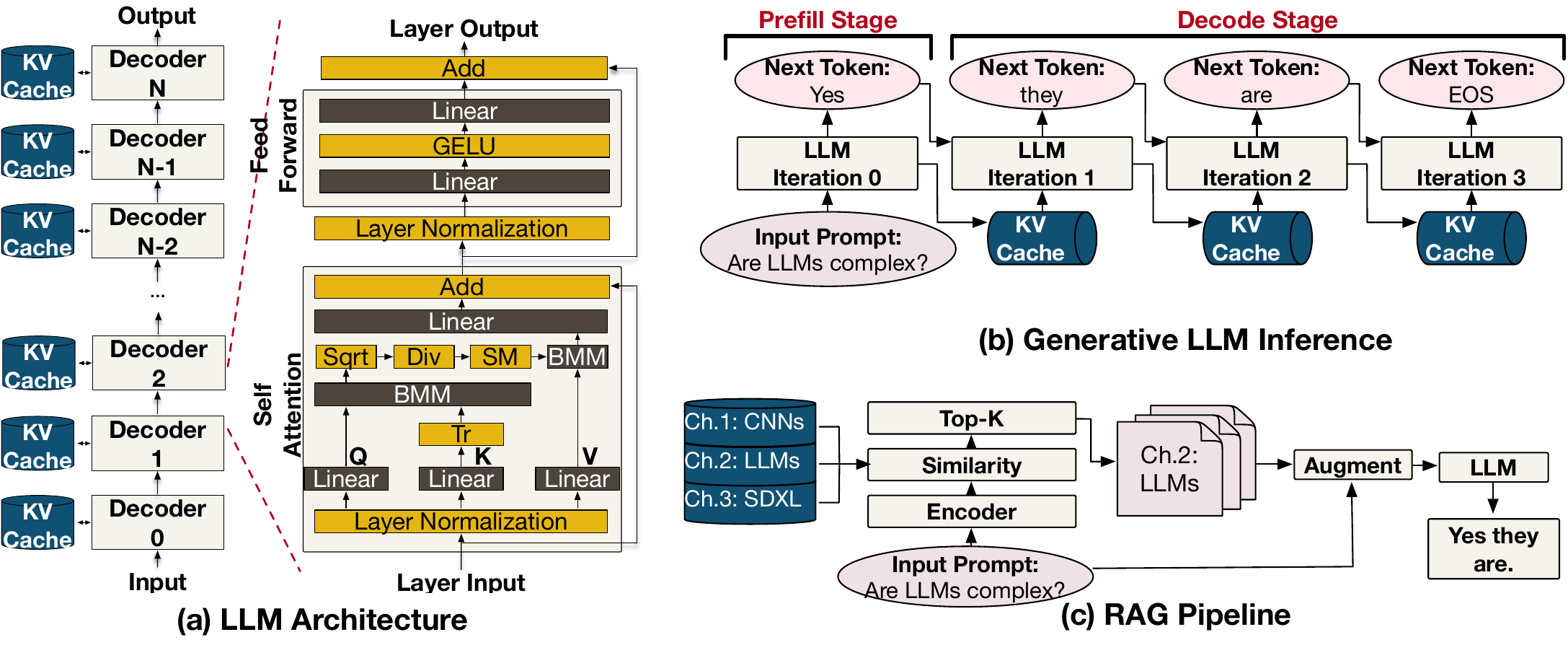}{ Overview of LLM architecture and generative inference.}{0.75}

State-of-the-art large language models~\cite{achiam2023gpt, grattafiori2024llama3, zhang2022opt} (LLMs) are build by cascading multiple transformer blocks as shown in~\autoref{fig:LLM-architecture} (a).
Each transformer~\cite{vaswani2017attention} block consists of a self-attention layer followed by a feed-forward layer. 
Given an input tensor $X$ of size $S \times D$, where S is the length of the input sequence and D is the hidden dimension of the LLM, a transformer block first normalizes the input tensor and feeds the normalized tensor to the self-attention layer. 
The self-attention layer multiplies the normalized tensor by three weight matrices ($W_{Q}, W_{K}, W_{V}$) each of size $D \times D$ to generate the query $Q$, key $K$, and value $V$ matrices of size $N \times D$.
Then, the self-attention layer computes the attention scores by applying $softmax(QK^T)V$, and forwards the resulting tensor to the feed-forward layer. 

Generative LLM inference consists of two main stages: \textit{prefill}, and \textit{decode} shown in~\autoref{fig:LLM-architecture} (b). 
During the prefill stage, the LLM generates the first output token and populates the KV cache. 
The KV cache stores the $K$ and $V$ matrices of every processed token at every layer of the model.
This common optimizations eliminates redundant $K$ and $V$ recomputation in the iterative decode stage. 
During the decode stage, the LLM auto-regressively generates new tokens. 
It uses the last generated token in the sequence and the KV cache entries of all the previous tokens to generate the next token. 

Recently, LLMs have been integrating retrieval augmented generation (RAG)~\cite{lewis2020retrieval,asai2023selfrag} to augment the input context with relevant information retrieved from a database at real-time. 
\autoref{fig:LLM-architecture} (c) highlights the main components of a RAG pipeline: the encoder and the generator.
% typically consist of an encoder and a \textit{generator}. 
% 
The RAG pipeline pipeline's encoder computes embeddings of the input prompt, and selects the top-K database entries that is the most similar to the input prompt.  
The input prompt is next augmented with the retrieved information and fed to the generator which is a LLM to generate the output response.
% 
% The retriever encodes the input prompt/sequence and computes the similarity with the encoded information in the given database. 
% %  
% The top-K database entries with the highest similarity are selected and used to augment the input prompt.
% % 
% The augmented input prompt is fed to the generator which is a generative LLM in our case. 

% 

\subsection{Real-Time Workloads}
\label{subsec:rt-models}
% \insertFigure{prelim-sim-overview}{Overview of the runtime simulation used to evaluate the workloads scenarios defined in~\autoref{tab:scenarios} \TODO{include choice of scheduling policy in inputs}}

\insertFigure{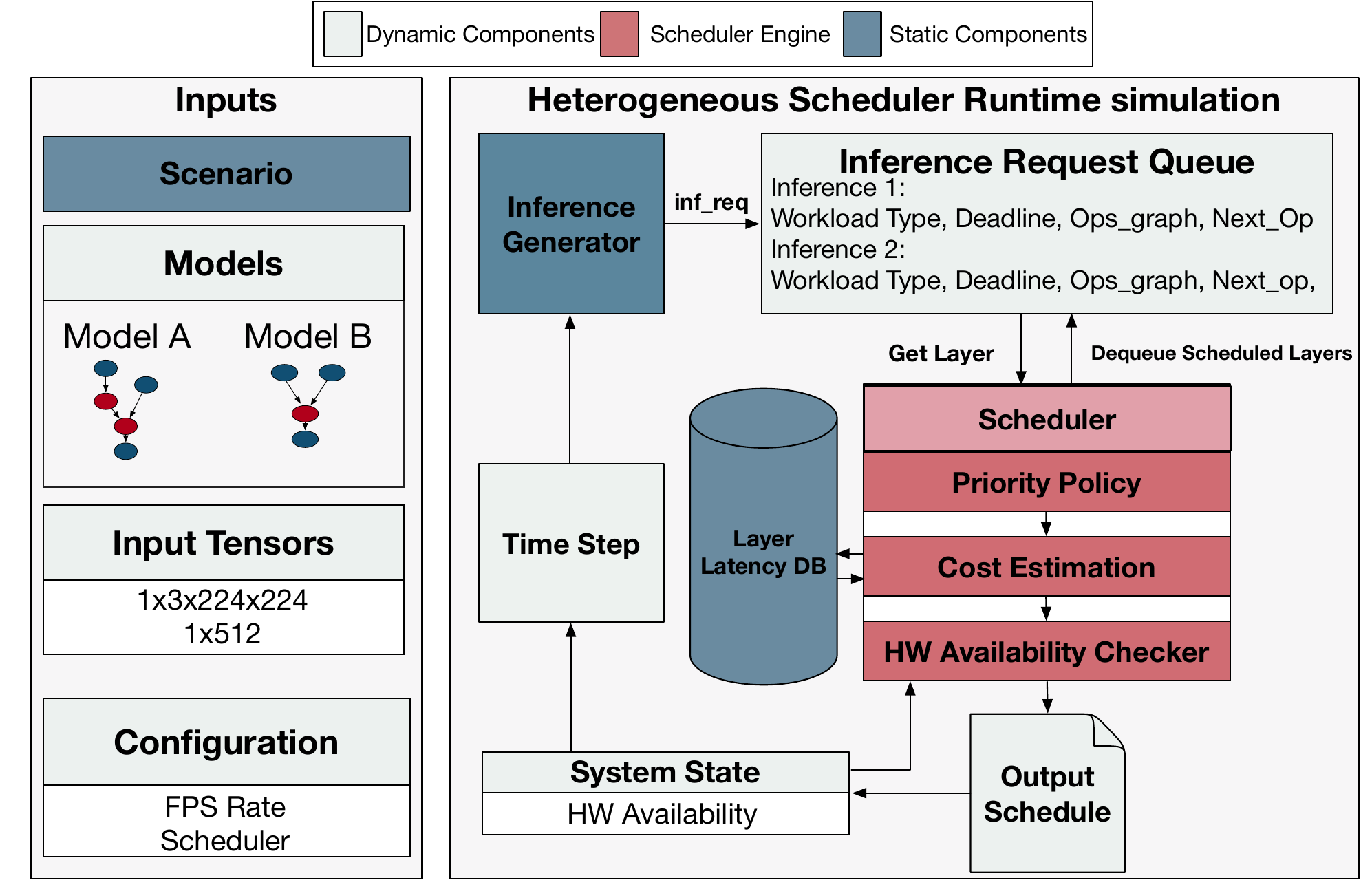}{Overview of the runtime simulation used to evaluate the workloads scenarios defined in~\autoref{tab:scenarios}}

The success of ML models across computer vision and speech recognition tasks lead to an increased adoption of these models in various real-time applications~\cite{kwon2023xrbench, zoomaiassistant, teams, gallotta2024llmgaming}.
Video conferencing and gaming applications often integrate ML based features to improve the real-time user experience~\cite{valevski2024diffusiongames, gallotta2024llmgaming,  zoomaiassistant, teams}.
For example, many video conferencing applications rely on segmentation DNNs to enable background blurring and customizations~\cite{teams, lugaresi2019mediapipe, lin2022efficient}.
These applications also use super-resolution networks~\cite{zhang2018SR,teamsSR,amd_fsr } to improve the user's quality of experience. 
This class of ML applications requires ML inference requests to be periodically serviced to provide a continuous and smooth user experience. 
Since the inference latency directly impact the user quality of experience, strict real-time processing requirements, quantified using the frame per second (FPS) processing rates, must be met.
\insertFigureShrink{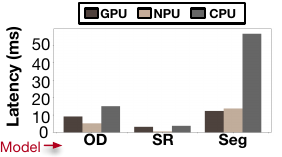}{The impact of the hardware platform on the inference latency of selected real-time models.}{0.8}

\insertFigure{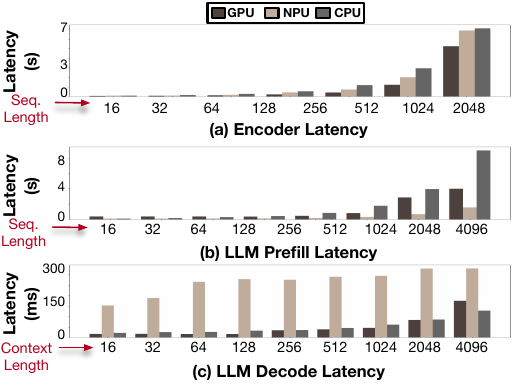}{Comparing the latency of the (a) encoder, (b) prefill, and (c) decode stages of the RAG pipeline used in scenario A.}

% ML models are being integrated into real-time applications such as video conferencing, gaming, etc. 
% %  
% Increase integration of ML in real time applications --> More models that target real-time applications. 
% RT models characteristics: Multi-modal, strict FPS requirements. 
% Example of RT models. 

% \begin{itemize}
%     \item Super Resolution: RCAN (RyzenAI SW - ECCV 2018)
%     \item Segmentation:  PointPainting (RyzenAI SW - CVPR 2020)
%     \item Speech to Text: 
% \end{itemize}

\subsection{Heterogeneous Systems}
The heterogeneity in ML workloads paired with the increased demand of high performance inference lead to an increased adoption of heterogeneous systems to meet the computational and memory requirements of the inference workloads. 
Heterogeneous systems typically incorporate two or more distinct compute units (CPU, GPU, NPU).

\betterparagraph{CPU-GPU Systems} CPU-GPU systems comprise of a CPU connected to a discrete GPU via a PCI Express (PCIe) link.
This type of heterogeneous systems is widely adopted across platforms from edge (e.g., laptops, workstations) to cloud devices.
GPUs are widely adopted for deep learning inference due to their high computational throughput. For instance, the AMD Radeon RX 7900 XTX offers up to 123 half precision TFLOPS~\cite{amd_radeon_rx7900xtx}. 
% 
% The CPUs and GPUs have each a separate physical memory, necessitating costly explicit CPU-GPU data movements.
% %  
% Extensive research and industrial efforts have optimized GPU inference performance, making them the de facto choice for high-performance workloads. 
% 
However, GPUs consume significant amounts of energy, which poses a major challenge for power-constrained edge deployments~\cite{rakka2025softmap}. 

\betterparagraph{Integrated Heterogeneous Systems} The increasing complexity and heterogeneity of machine learning workloads and the efforts to increase accessibility to GenAI have driven the development of tightly integrated heterogeneous platforms. 
Modern System-on-Chip (SoC) architectures, such as Apple’s M-series~\cite{apple_m3_chips}, AMD Ryzen AI~\cite{amd_ryzen_ai}, and Qualcomm Snapdragon~\cite{qualcomm_snapdragon_pcs}, tightly integrate CPUs, NPUs, and GPUs on a single package. 
The tight integration of multiple compute engines unlocks potential opportunities for additional optimizations.
%  
% CPU-GPU: Integrated cpu-gpu like ryzenAI 
% moving to integrated memory 

% In the accelerator domain, using different accelerators (heterogeneous) 

% Scheduling on the heterogeneous System 

% Difference between NPUs and GPUs

\begin{comment}
\subsubsection{Diffusion Models}
% \label{subsubsec:diffusion}

% Diffusion models generate images through an iterative process. 
% % 
% They start from a random noisy input and gradually transform it into a coherent image guided by an input text, by running the noisy input through a denoising backbone for multiple iterations. 
% % 
% Diffusion models' denoising backbone is based on an encoder-decoder modified U-Net architecture~\cite{ronneberger2015unet}. 
% % 
% The modified U-Net layers alternate between ResNet~\cite{he2016deep}, self-attention~\cite{vaswani2017attention}, and cross-attention~\cite{chen2021crossvit} blocks.
% %
\end{comment}
\section{GenAI in Real-Time Applications}
\label{sec:genai_rt}
\rtgen workloads represent an emerging class of ML inference tasks that combine the challenges of generative and real-time models in a scenario-driven, multi-model setting.
% \rtgen workloads represent an emerging ML inference task that combines the challenges of generative and real-time models in a scenario-driven, multi-model setting.
We outline the characteristics of multi-model \rtgen workloads and define four representative scenarios,summarized in~\autoref{tab:scenarios}, inspired by real applications.

% We present a set of multi-model workload scenarios that integrate RT and Gen tasks, designed to reflect emerging AI applications on modern AI PCs. 
% % 
% These scenarios capture the diversity and concurrency found in real-world use cases. 
% % 
% Each scenario is defined by the models involved, target frame rates (FPS), and their dynamic execution characteristics, as summarized in Table\ref{tab:scenarios}.
% 
 % \insertFigure{challenges} {Running RT and GenAI workloads results in dynamic resource availability. \TODO{Shrink Figure, Update models} }
 % Show multiple scenarios to highlight the challenges in the scheduling problem: wait to run on preferred backend, or run on inefficient backends, Heterogeneity in workload (diverse performance across workloads) each subplot should reflect a challenge

% Scenario A: AI Assistant
%   RAG Pipeline 
%       Encoder + LLM 
% Scenario B: Debugging Call 
%   RAG Pipeline (link) 
%   Speech to text + RAG Pipeline (link) 
% Scenario C: Design Brainstorming Video Conference / Online Media Content Creation 
%   Background Blurring: Semantic Segmentation 
%   Real Time Summaries: Smart Speech + RAG Pipeline (link)
%   Design Brainstorming: Diffusion + Super Resolution (link) (link) (link)
% Scenario D: Gaming and Streaming Assistant
%   Super Resolution 
%   LLM (link)

\begin{table*}
\caption{Unit models used to construct the evaluated workload scenarios. TTFT, TPT, and E2E refer to time to first token, time per token, and end-to-end latency respectively.}

\scriptsize
\centering

\begin{tabular}{|l|l|l|l|}
\hline
\multicolumn{1}{|c|}{\multirow{1}{*}{\textbf{Application}}} 
& \multicolumn{1}{c|}{\multirow{1}{*}{\textbf{Task}}} 
& \multicolumn{1}{c|}{\multirow{1}{*}{\textbf{Models}}}
& \multicolumn{1}{c|}{\multirow{1}{*}{\textbf{Task Perf. Metric}}} \\ 
\hline

\multicolumn{1}{|c|}{\multirow{3}{*}{\textbf{Generative Agent}}} 
& \multicolumn{1}{c|}{\multirow{2}{*}{\textbf{RAG Pipeline (RAG)}}} 
& \multicolumn{1}{c|}{\multirow{1}{*}{\textbf{GTE Encoder~\cite{li2023gte}}}}
& \multicolumn{1}{c|}{\multirow{3}{*}{\textbf{TTFT, TPT, E2E}}} \\ 
\cline{3-3}
& & \multicolumn{1}{c|}{\multirow{1}{*}{\textbf{Llama3-1B~\cite{grattafiori2024llama3}}}}
& \\ 
\cline{2-3}

& 
\multicolumn{1}{c|}{\multirow{1}{*}{\textbf{LLM}}} 
& \multicolumn{1}{c|}{\multirow{1}{*}{\textbf{Llama3-1B~\cite{grattafiori2024llama3}}}}
% 
% & \multicolumn{1}{c|}{\multirow{1}{*}{\textbf{TTFT, TPT, E2E}}} \\ 
& \\
\hline
\multicolumn{1}{|c|}{\multirow{1}{*}{\textbf{Background Blurring}}} 
& \multicolumn{1}{c|}{\multirow{1}{*}{\textbf{Segmentation (Seg)}}} 
& \multicolumn{1}{c|}{\multirow{1}{*}{\textbf{PointPainting~\cite{vora2020pointpainting}}}}
& \multicolumn{1}{c|}{\multirow{3}{*}{\textbf{Frame Drop Rate}}} \\ 
\cline{1-3}
\multicolumn{1}{|c|}{\multirow{1}{*}{\textbf{Resolution Enhancement}} }
& \multicolumn{1}{c|}{\multirow{1}{*}{\textbf{Super Resolution (SR)}}} 
& \multicolumn{1}{c|}{\multirow{1}{*}{\textbf{RCAN ~\cite{zhang2018SR}}}}
% 
% & \multicolumn{1}{c|}{\multirow{1}{*}{\textbf{Frame Drop Rate}}} \\ 
& \\
\cline{1-3}
\multicolumn{1}{|c|}{\multirow{1}{*}{\textbf{Hand Gesture Detection}} }
& \multicolumn{1}{c|}{\multirow{1}{*}{\textbf{Object Detection (OD))}}} 
& \multicolumn{1}{c|}{\multirow{1}{*}{\textbf{YOLOv3 ~\cite{redmon2018yolov3}}}}
% 
% & \multicolumn{1}{c|}{\multirow{1}{*}{\textbf{Frame Drop Rate}}} \\ 
& \\
\hline

\end{tabular}
\label{tab:unit_models}
\end{table*}

\subsection{\rtgen Characteristics}

\rtgen workloads pose unique challenges due to the integration of generative LLMs with real-time, single- and multi-model inference workloads.
Therefore, we define the following characteristics of emerging \rtgen scenarios: dynamic input shapes,  real-time processing, multi-model execution, and model heterogeneity.
\betterparagraph{Dynamic Input Shapes} Generative LLMs in \rtgen scenarios process tensors with dynamic shapes, depending on the input sequence length and execution stage.
% 
% The shape of the tensors processed by the LLM depend on the input sequence length and the execution stage. 
% 
During the prefill stage, the tensors have a $S \times D$ shape, $S$ being the input sequence length and $D$ the model's hidden dimension. 
During the decode stage, the LLM auto-regressively generated a single token at a time by using inputs of shape $1 \times D$ in every decode iteration.
The dynamically varying input sequence lengths of a generative LLM between the prefill and decode stages alter the LLM's preferred backend. 
% These variations can alter a model’s runtime behavior and can shift its preferred backend. 
% 
For instance, as shown in~\autoref{fig:llm-profiling}, the LLM prefers the NPU during the compute-intensive prefill stage, but chooses the GPU for the memory- and bandwidth-intensive decode stage. 
% an input prompt that runs efficiently on the NPU at large sequence lengths may perform better on a GPU at shorter sequence as illustrated in~\autoref{fig:llm-profiling}. 
% 
As a result, static ahead-of-time (AoT) input-unaware scheduling will fail to capture input dynamicity leading to sub-optimal performance.

\betterparagraph{Real-Time Constraints} \rtgen applications include real-time tasks with strict frame rate requirements as shown in ~\autoref{tab:scenarios}.
Meeting these FPS targets is critical to maintaining quality of experience in applications such as real-time background blurring in video conferencing and resolution enhancement in gaming.
Frame drops caused by missed deadlines directly degrade the user's overall experience.

% These requirements are critical for preserving the quality of experience in applications such as real-time background blurring in video conferencing, resolution enhancement in gaming, etc.
% 
% Failing to meet the target FPS rates results in frame drops leading to a degradation of the application performance and the overall user experience.

\betterparagraph{Multi-Model Execution} \rtgen scenarios require the execution of multiple models, often concurrently and in a cascaded manner.
% /
%
This concurrency can lead to dynamic backend availability, introducing significant challenges for scheduling—especially in the presence of models with dynamic tensor shapes.
% The concurrent execution of multiple models can lead to dynamic backend availability, which imposes additional challenges on scheduling, especially when processing models with dynamic tensor shapes. 
%  

\betterparagraph{Model Heterogeneity} \rtgen workloads deploy models from different task domains with diverse architectures. 
For example, scenario B in~\autoref{tab:scenarios} consists of a RAG and a super-resolution (SR) model. 
The RAG's encoder and generator have both transformer based architectures, while the SR network is a convolutional neural network (CNN). 
This architectural diversity further complicates scheduling and backend selection.
% execution of multiple models, each with different hardware resource requirements.
% 
% The presence of multiple active models leads to dynamic backend availability.
% %  
% This dynamic state of the system complicates scheduling, especially when different models prefer similar hardware backends.

\subsection{Usage Scenarios}
% \input{Tables/tab_scenarios}
% Scenario A: AI Assistant
%   LLM Pipeline 
%       Encoder + LLM 
% Scenario B: Debugging Call 
%   LLM Pipeline (link) 
%   Speech to text + LLM Pipeline (link) 
% Scenario C: Design Brainstorming Video Conference / Online Media Content Creation 
%   Background Blurring: Semantic Segmentation 
%   Real Time Summaries: Smart Speech + LLM Pipeline (link)
%   Design Brainstorming: Diffusion + Super Resolution (link) (link) (link)
% Scenario D: Gaming and Streaming Assistant
%   Super Resolution 
%   LLM (link)

\begin{table*}
\caption{Description of the real-time - generative workload scenarios and their respective target processing rates in FPS.}
\scriptsize 
\center
\begin{tabular}{|l|l|l|l|l|l|l|l|l|l|}

\hline
\multicolumn{1}{|c|}{\multirow{3}{*}{\textbf{ID}}}
& \multicolumn{1}{|c|}{\multirow{3}{*}{\textbf{Scenario}}}
& \multicolumn{1}{|c|}{\multirow{3}{*}{\textbf{Tasks}}}
& \multicolumn{4}{|c|}{\multirow{1}{*}{\textbf{Characteristics}}}
& \multicolumn{1}{|c|}{\multirow{3}{*}{\textbf{Scenario Description}}} \\
\cline{4-7}

&
&
& \multicolumn{1}{|c|}{\multirow{1}{*}{\textbf{Real-Time}}} 
&\multicolumn{1}{|c|}{\multirow{1}{*}{\textbf{Dynamic}}} 
&\multicolumn{1}{|c|}{\multirow{1}{*}{\textbf{Multi-}}}

&\multicolumn{1}{|c|}{\multirow{1}{*}{\textbf{Model-}}}
& \\
&
&
&\multicolumn{1}{|c|}{\multirow{1}{*}{\textbf{Processing}}}
&\multicolumn{1}{|c|}{\multirow{1}{*}{\textbf{Shapes}}}
&\multicolumn{1}{|c|}{\multirow{1}{*}{\textbf{Model Exec.}}}
&\multicolumn{1}{|c|}{\multirow{1}{*}{\textbf{Heterogeneity}}}
& \\
\hline

\multicolumn{1}{|c|}{\multirow{1}{*}{\textbf{A}}} 
& \multicolumn{1}{|c|}{\multirow{1}{*}{\textbf{AI Chatbot Assistant}}}
&\multicolumn{1}{|c|}{\multirow{1}{*}{\textbf{RAG}}}
&\multicolumn{1}{|c|}{\multirow{1}{*}{\textbf{ }}}
&\multicolumn{1}{|c|}{\multirow{1}{*}{\textbf{\checkmark}}}
&\multicolumn{1}{|c|}{\multirow{1}{*}{\textbf{ }}}
&\multicolumn{1}{|c|}{\multirow{1}{*}{\textbf{\checkmark}}}
&\multicolumn{1}{|c|}{\multirow{1}{*}{RAG-based chatbot to answer the users prompts}} \\
\hline
% 
% Scenario B
\multicolumn{1}{|c|}{\multirow{2}{*}{\textbf{B}}} 
& \multicolumn{1}{|c|}{\multirow{2}{*}{\textbf{Gaming I}}}
& \multicolumn{1}{|c|}{\multirow{1}{*}{\textbf{LLM}}}
&\multicolumn{1}{|c|}{\multirow{1}{*}{\textbf{ }}}
&\multicolumn{1}{|c|}{\multirow{2}{*}{\textbf{\checkmark}}}

&\multicolumn{1}{|c|}{\multirow{2}{*}{\textbf{\checkmark}}}
&\multicolumn{1}{|c|}{\multirow{2}{*}{\textbf{\checkmark}}}
&\multicolumn{1}{|c|}{\multirow{2}{*}{LLM-based non-playable characters}} \\
\cline{3-3}
\cline{4-4}

&
&\multicolumn{1}{|c|}{\multirow{1}{*}{\textbf{SR}}}
&\multicolumn{1}{|c|}{\multirow{1}{*}{\textbf{120 FPS}}}
&
&
&
& \\
\hline 
% 
% Scenario C

\multicolumn{1}{|c|}{\multirow{4}{*}{\textbf{C}}} 
& \multicolumn{1}{|c|}{\multirow{4}{*}{\textbf{Gaming II}}}
& \multicolumn{1}{|c|}{\multirow{1}{*}{\textbf{RAG}}}
&\multicolumn{1}{|c|}{\multirow{1}{*}{\textbf{ }}}
&\multicolumn{1}{|c|}{\multirow{4}{*}{\textbf{\checkmark}}}
&\multicolumn{1}{|c|}{\multirow{4}{*}{\textbf{\checkmark}}}
&\multicolumn{1}{|c|}{\multirow{4}{*}{\textbf{\checkmark}}}
&\multicolumn{1}{|c|}{\multirow{4}{*}{RAG-based game streaming assistant}} \\
\cline{3-3}
\cline{4-4}
&
& \multicolumn{1}{|c|}{\multirow{1}{*}{\textbf{SR-120}}}
&\multicolumn{1}{|c|}{\multirow{1}{*}{\textbf{120 FPS}}}
&
&
&
& \\
\cline{3-3}
\cline{4-4}
&
& \multicolumn{1}{|c|}{\multirow{1}{*}{\textbf{SR-60}}}
&\multicolumn{1}{|c|}{\multirow{1}{*}{\textbf{60 FPS}}}
&
&
&
& \\
\cline{3-3}
\cline{4-4}
&
& \multicolumn{1}{|c|}{\multirow{1}{*}{\textbf{Seg}}}
&\multicolumn{1}{|c|}{\multirow{1}{*}{\textbf{60 FPS}}}
&
&
&
& \\
\hline
% 
% 
% Scenario D
\multicolumn{1}{|c|}{\multirow{4}{*}{\textbf{D}}} 
& \multicolumn{1}{|c|}{\multirow{4}{*}{\textbf{Video Conference}}}
& \multicolumn{1}{|c|}{\multirow{1}{*}{\textbf{RAG}}}
&\multicolumn{1}{|c|}{\multirow{1}{*}{\textbf{ }}}
&\multicolumn{1}{|c|}{\multirow{4}{*}{\textbf{\checkmark}}}
&\multicolumn{1}{|c|}{\multirow{4}{*}{\textbf{\checkmark}}}
&\multicolumn{1}{|c|}{\multirow{4}{*}{\textbf{\checkmark}}}
&\multicolumn{1}{|c|}{\multirow{4}{*}{Video conference with RAG-based chatbot assistant}} \\
\cline{3-3}
\cline{4-4}
&
& \multicolumn{1}{|c|}{\multirow{1}{*}{\textbf{SR}}}
&\multicolumn{1}{|c|}{\multirow{1}{*}{\textbf{60 FPS}}}
&
&
&
& \\
\cline{3-3}
\cline{4-4}
&
& \multicolumn{1}{|c|}{\multirow{1}{*}{\textbf{Seg}}}
&\multicolumn{1}{|c|}{\multirow{1}{*}{\textbf{60 FPS}}}
&
&
&
& \\
\cline{3-3}
\cline{4-4}
&
& \multicolumn{1}{|c|}{\multirow{1}{*}{\textbf{OD}}}
&\multicolumn{1}{|c|}{\multirow{1}{*}{\textbf{60 FPS}}}
&
&
&
& \\
\hline

\end{tabular}
\label{tab:scenarios}
\end{table*}

We use models from ~\autoref{tab:unit_models} to define four representative scenarios highlighted in ~\autoref{tab:scenarios}, each inspired by practical edge AI applications.

\betterparagraph{Scenario A: AI Assistant} This scenario involves a RAG pipeline for text-based prompt answering. 
This scenario does not have real-time constraints, but it exhibits input dynamicity caused by the varying input prompt length of the RAG's LLM. 

\betterparagraph{Scenario B: Gaming I} This is a multi-model scenario that simulates a gaming scenario where the user interacts with a smart non-playable character~\cite{park2023generative, gallotta2024llmgaming}. 
This scenario consist of a generative LLM and a SR network.
The generative LLM pipeline drives the non-playable characters to converse with the user and  answer their questions, and the real-time SR model improves the resolution of the game~\cite{amd_fsr,zhang2018SR}.  

\betterparagraph{Scenario C: Gaming II} This scenario represents a live-stream gaming session assistant.
A segmentation model runs for background blurring and manipulation~\cite{vora2020pointpainting, teams, }, while 2 super-resolution models runs to enhance the resolutions of the game ~\cite{amd_fsr,zhang2018SR} and the streamed video~\cite{teamsSR}. 
A RAG based AI Assistant answers the user's questions during the streaming process for a smoother experience~\cite{streamlabs_ai_assistant}.
% LLM powers smart non-playable-characters for a better interactive experience, adding a generative component to the real-time pipeline.

\betterparagraph{Scenario D: Video Conference} The scenario describes a video conference call powered by a smart AI assistant. assistant~\cite{microsoft_copilot_orgs, zoomaiassistant}. 
Scenario D deploys a RAG based assistant and includes a real-time segmentation model to support background blurring ~\cite{teams,zoom_blur_background}, a real-time object detection (OD) model for gesture recognition~\cite{zoom_gesture_recognition}, and a SR network for enhancing the quality of the video. 
% 

% \TODO{Separate section after background}
% \TODO{Table showing workload scenarios, showing models, model concurrency, model cascading, target FPS,}
% \begin{itemize}
%     \item {Challenges in LLM and RAG: Dynamic Input Length -> Varying Compute-to-Memory ratio, Varying HW preference, Inference Memory Footprint }
%     % \item {Challenges in Diffusion: Memory Footprint, HW preference}
%     \item {Challenges in Real-Time Models: Target FPS}
% \end{itemize}
% \TODO{Discuss the potential implication of the RAG pipeline: Additional overheads to encode+compute-similarity, Dynamic inputs due to varying size input prompts}. 
% \TODO{The challenges when combining both workloads: Dynamic Resource Availability}
% \TODO{Figure showing the resource utilization/availability over time}
\section{Methodology}
\label{sec:methodology}

\subsection{Profiling Methodology}

\begin{table*}[]
\centering
% \scriptsize
\caption{Hardware platform configurations used for case studies.}
\begin{tabular}{|l|l|l|l|l|l|l|l|}
\hline
\multicolumn{1}{|c|}{\multirow{2}{*}{\textbf{ID}}} 

& \multicolumn{2}{c|}{\multirow{1}{*} {\textbf{CPU}}}             
& \multicolumn{2}{c|}{\multirow{1}{*}{\textbf{NPU}}}
& \multicolumn{2}{c|}{\multirow{1}{*}{\textbf{GPU}}}
\\ 
\cline{2-7}
& \multicolumn{1}{c|}{\textbf{Device}}   
& \multicolumn{1}{c|}{\textbf{Memory}}             

& \multicolumn{1}{c|}{\textbf{Device}}             
% & \multicolumn{1}{c|}{\textbf{Mem.}}
& \multicolumn{1}{c|}{\textbf{TOPS}} 
& \multicolumn{1}{c|}{\textbf{Device}}             
% & \multicolumn{1}{c|}{\textbf{Mem.}}
& \multicolumn{1}{c|}{\textbf{TOPS}} \\ 
\hline
A                                   
& \multicolumn{1}{l|}{AMD Ryzen AI 9 365}     
& \multicolumn{1}{l|}{24 GB}                           

& \multicolumn{1}{l|}{XDNA 2}      
% & \multicolumn{1}{r|} {11.6 GB}
& \multicolumn{1}{r|} {50} 
& \multicolumn{1}{l|}{AMD Radeon 880M}      
% & \multicolumn{1}{r|} {12 GB}
& \multicolumn{1}{r|} {16} \\ 
\hline
%%
% B                            
% & \multicolumn{1}{l|}{AMD Ryzen 9 7940HS}                         
% & \multicolumn{1}{l|}{XDNA}  
% & \multicolumn{1}{r|} {15.6 GB}
% & \multicolumn{1}{r|} {10} 
% & \multicolumn{1}{l|}{AMD Radeon 780M}  
% & \multicolumn{1}{r|} {16.1 GB}
% & \multicolumn{1}{r|} {660*} \\ 
% \hline
%%
\end{tabular}
% \vspace{-5mm}
\label{tab:experiment_settings}
\end{table*}
To profile the models in~\autoref{tab:unit_models}, we utilize the AMD open source RyzenAI SW~\cite{amd_ryzenai_sw}.
The RyzenAI SW flow enables deploying models on both the NPU and the GPU of Ryzen AI processor described in ~\autoref{tab:experiment_settings}, by leveraging the ONNX Runtime (ORT)~\cite{onnxruntime} and the ONNX runtime generative AI (OGA)~\cite{oga} inference frameworks.
The RyzenAI SW toolchain utilize the VitisAI execution provider~\cite{onnx_vitis_ai} for NPU deployment and the the DirectML execution provider~\cite{onnx_directml} for GPU deploymnent. 

To profile the LLM, the model is first quantized using the AWQ post-quantization~\cite{lin2024awq} scheme to reduce the model's weights to a 4 bit INT4 precision.
Then, the quantized LLM graph is passed to the OGA runtime which partitions the LLM's graph and maps it to the chosen backend (NPU or GPU). 
Afterwards we run the LLM for multiple input and output sequence lengths and record the latency of both the prefill and the decode stage. 

Similarly, we quantize the RAG's encoder to BFloat16 and Float16 for NPU and GPU deployment respectively. We also quantize the SR, Seg, and OD networks to INT8. 
We then use ORT with the respective execution provider to run the quantized models and collect their inference latency. 
We use the collected latency profiles to drive~\autoref{tab:scenarios}'s scenarios simulation experiments. 

% We profile the models in~\autoref{tab:unit_models} on the AMD Ryzen AI processor listed in~\autoref{tab:experiment_settings}. 
% %  
% We use AMD's open source RyzenAI SW~\cite{amd_ryzenai_sw} and ONNX Runtime (ORT)~\cite{onnxruntime} to run the model inferences. 
% %
% We utilize the VitisAI execution provider~\cite{onnx_vitis_ai} and the DirectML execution provider~\cite{onnx_directml} to deploy the models on the NPU and the GPU respectively. 
% % 
% We measure the inference latency and the inference latency breakdown using RyzenAI and ORT specific profilers.

\subsection{Runtime Simulator}
\label{subsec:runtime_sim}
To evaluate the impact of scheduling on the performance of the emerging \rtgen workloads, we implement a Python-level simulator highlighted in~\autoref{fig:prelim-sim-overview} to simulate the runtime of the different scenarios defined in~\autoref{tab:scenarios} under the five scheduling policies described in~\autoref{tab:schedulers}. 
% %  
% \autoref{fig:prelim-sim-overview} shows an overview of the simulator used to perform the experiments. 
%  
The simulator inputs consist of the scenario's models, their input tensors, the real-time requirements (e.g. FPS), as well as a scheduling policy. 
% The simulator takes as input the scenario's models, their input tensors, and their respective real-time requirements (e.g. FPS).
% %  
Given those models, the inference generator issues inference requests based to the inference request queue based on each model FPS. 
Next, the scheduler module selects a layer or a model to be scheduled based on the system state, the cost of running the layer/model, and the selected scheduling policy. 
To estimate the deployment cost of every layer/model on a given backend, the scheduler relies on a latency database constructed using the latencies of the profiled models. 

\subsection{Scheduling Methodology}
\label{subsec:scheduling}

We evaluate the workload scenarios under the five scheduling policies listed in~\autoref{tab:schedulers}. To enable a structured analysis of their trade-offs, we introduce the following taxonomy to captures the key features relevant to scheduling \rtgen workloads:

\begin{itemize}
    \item \textbf{Deadline Awareness:} A scheduling policy is considered deadline-aware if it incorporates real-time constraints, such as frame rate deadlines,into its decision-making process.

    \item \textbf{Dynamic Hardware Selection:} A policy supports dynamic hardware selection if it determines the execution backend (NPU or GPU) at runtime, based on current system load and availability. This contrasts with ahead-of-time (AoT) policies that statically assign backends regardless of runtime conditions.

    \item \textbf{Heterogeneity Awareness:} A heterogeneous-aware policy is able to utilize multiple backends concurrently, taking advantage of both NPU and GPU resources to maximize parallelism and throughput.

    \item \textbf{GenAI Awareness:} GenAI-aware policies recognize the unique latency characteristics of generative models (e.g., LLMs) and ensures that long-latency models are not starved or preempted unnecessarily.
\end{itemize}

Each of the five scheduling policies differs in how it aligns with these taxonomy dimensions:

\betterparagraph{First-Come First-Serve AoT (FCFS-AOT)} This policy schedules the next layer from the oldest inference request in the queue. The backend is selected statically ahead of time based on latency profiling. FCFS-AOT does not account for real-time deadlines, does not adapt to runtime hardware availability, and treats all models equally without regard to generative workloads.

\betterparagraph{First-Come First-Serve Dynamic (FCFS-DYN):} Like FCFS-AOT, this policy schedules requests in arrival order, but it dynamically selects the backend at runtime based on current availability. It still lacks deadline awareness and does not distinguish between real-time and generative workloads.

\betterparagraph{Earliest Deadline First AoT (EDF-AOT):} This policy prioritizes inference requests based on their deadlines, scheduling the layer from the request with the closest deadline. However, backend selection remains static and fixed prior to execution. While EDF-AOT introduces deadline awareness, it lacks both dynamic backend selection and GenAI-specific treatment.

\betterparagraph{Earliest Deadline First Dynamic (EDF-DYN):} EDF-DYN improves upon EDF-AOT by choosing the execution backend dynamically at runtime. It retains the same deadline-aware behavior but adds runtime flexibility in hardware selection.

\betterparagraph{First Token First (FTF):}FTF extends EDF-DYN with GenAI-specific enhancements. It prioritizes LLMs' TTFT by treating the first token as a high-priority deadline, mitigating starvation during the compute-intensive prefill phase. Once the TTFT is served, FTF falls back to EDF-DYN behavior, since the decode stage latencies are more akin to the other non-generative models. This approach balances the stringent latency demands of real-time models with the long and stage-dependent latency of LLMs.

\begin{table}
\caption{Evaluated Scheduling Policies.}
\scriptsize 
\center
\begin{tabular}{|l|l|l|l|l|l|l|l|l|l|}

\hline
\multicolumn{1}{|c|}{\multirow{2}{*}{\textbf{Scheduler}}}
&\multicolumn{1}{|c|}{\multirow{1}{*}{\textbf{Deadline}}}
&\multicolumn{1}{|c|}{\multirow{1}{*}{\textbf{Dynamic HW}}}
&\multicolumn{1}{|c|}{\multirow{1}{*}{\textbf{Heterogeneity }}}
&\multicolumn{1}{|c|}{\multirow{1}{*}{\textbf{GenAI}}}\\
% 
% &\multicolumn{1}{|c|}{\multirow{1}{*}{\textbf{Backends}}}
% 
&\multicolumn{1}{|c|}{\multirow{1}{*}{\textbf{Aware}}}
&\multicolumn{1}{|c|}{\multirow{1}{*}{\textbf{Selection}}}
&\multicolumn{1}{|c|}{\multirow{1}{*}{\textbf{Aware}}} 
&\multicolumn{1}{|c|}{\multirow{1}{*}{\textbf{Aware}}}\\
\hline
\multicolumn{1}{|c|}{\multirow{1}{*}{\textbf{EDF-AOT}}}
% 
% &\multicolumn{1}{|c|}{\multirow{1}{*}{\textbf{1}}}
% 
&\multicolumn{1}{|c|}{\multirow{1}{*}{\textbf{\checkmark}}}
&\multicolumn{1}{|c|}{\multirow{1}{*}{\textbf{\xmark}}}
&\multicolumn{1}{|c|}{\multirow{1}{*}{\textbf{\checkmark }}} 
&\multicolumn{1}{|c|}{\multirow{1}{*}{\textbf{\xmark }}} \\
% \cline{2-6}
% 
% 
% \multicolumn{1}{|c|}{\multirow{1}{*}{\textbf{EDF-Dynamic}}}
% 
% &\multicolumn{1}{|c|}{\multirow{1}{*}{\textbf{2}}}
% 
% &\multicolumn{1}{|c|}{\multirow{1}{*}{\textbf{\checkmark}}}
% % 
% &\multicolumn{1}{|c|}{\multirow{1}{*}{\textbf{\xmark}}}
% % 
% &\multicolumn{1}{|c|}{\multirow{1}{*}{\textbf{\checkmark }}} % 
% &\multicolumn{1}{|c|}{\multirow{1}{*}{\textbf{\xmark }}} \\
\hline

\multicolumn{1}{|c|}{\multirow{1}{*}{\textbf{EDF-Dyn}}}
% 
% &\multicolumn{1}{|c|}{\multirow{1}{*}{\textbf{1}}}
% 
&\multicolumn{1}{|c|}{\multirow{1}{*}{\textbf{\checkmark}}}
&\multicolumn{1}{|c|}{\multirow{1}{*}{\textbf{\checkmark}}}
&\multicolumn{1}{|c|}{\multirow{1}{*}{\textbf{\checkmark }}} % 
&\multicolumn{1}{|c|}{\multirow{1}{*}{\textbf{\xmark }}} \\
% \cline{2-6}
% 
% \multicolumn{1}{|c|}{\multirow{1}{*}{\textbf{FCFS-Dynamic}}}
% 
% &\multicolumn{1}{|c|}{\multirow{1}{*}{\textbf{2}}}
% 
% &\multicolumn{1}{|c|}{\multirow{1}{*}{\textbf{\checkmark}}}
% % 
% &\multicolumn{1}{|c|}{\multirow{1}{*}{\textbf{\checkmark}}}
% % 
% &\multicolumn{1}{|c|}{\multirow{1}{*}{\textbf{\checkmark}}} % 
% &\multicolumn{1}{|c|}{\multirow{1}{*}{\textbf{\xmark}}} \\
\hline

\multicolumn{1}{|c|}{\multirow{1}{*}{\textbf{FCFS-AOT}}}
% 
% &\multicolumn{1}{|c|}{\multirow{1}{*}{\textbf{1}}}
% 
&\multicolumn{1}{|c|}{\multirow{1}{*}{\textbf{\xmark}}}
&\multicolumn{1}{|c|}{\multirow{1}{*}{\textbf{\xmark}}}
&\multicolumn{1}{|c|}{\multirow{1}{*}{\textbf{\checkmark }}}
&\multicolumn{1}{|c|}{\multirow{1}{*}{\textbf{\xmark }}} \\
% \cline{2-6}

% \multicolumn{1}{|c|}{\multirow{1}{*}{\textbf{EDF-Dynamic-hete}}}
% 
% &\multicolumn{1}{|c|}{\multirow{1}{*}{\textbf{2}}}
% 
% &\multicolumn{1}{|c|}{\multirow{1}{*}{\textbf{\xmark}}}
% % 
% &\multicolumn{1}{|c|}{\multirow{1}{*}{\textbf{\xmark}}}
% % 
% &\multicolumn{1}{|c|}{\multirow{1}{*}{\textbf{\checkmark }}} 
% % 
% &\multicolumn{1}{|c|}{\multirow{1}{*}{\textbf{\xmark }}} \\
\hline

\multicolumn{1}{|c|}{\multirow{1}{*}{\textbf{FCFS-Dyn}}}
% 
% 
% &\multicolumn{1}{|c|}{\multirow{1}{*}{\textbf{1}}}
% 
&\multicolumn{1}{|c|}{\multirow{1}{*}{\textbf{\xmark}}}
&\multicolumn{1}{|c|}{\multirow{1}{*}{\textbf{\checkmark}}}
&\multicolumn{1}{|c|}{\multirow{1}{*}{\textbf{\checkmark }}} 
&\multicolumn{1}{|c|}{\multirow{1}{*}{\textbf{\xmark }}} \\
% \cline{2-6}

% \multicolumn{1}{|c|}{\multirow{1}{*}{\textbf{FCFS-Dynamic-hete}}}
% 
% 
% &\multicolumn{1}{|c|}{\multirow{1}{*}{\textbf{2}}}
% 
% &\multicolumn{1}{|c|}{\multirow{1}{*}{\textbf{\xmark}}}
% % 
% &\multicolumn{1}{|c|}{\multirow{1}{*}{\textbf{\checkmark}}}
% % 
% &\multicolumn{1}{|c|}{\multirow{1}{*}{\textbf{\checkmark }}}
% % 
% &\multicolumn{1}{|c|}{\multirow{1}{*}{\textbf{\xmark }}} \\
\hline

\multicolumn{1}{|c|}{\multirow{1}{*}{\textbf{FTF}}}
% 
% 
% &\multicolumn{1}{|c|}{\multirow{1}{*}{\textbf{1}}}
% 
&\multicolumn{1}{|c|}{\multirow{1}{*}{\textbf{\checkmark}}}
&\multicolumn{1}{|c|}{\multirow{1}{*}{\textbf{\checkmark}}}
&\multicolumn{1}{|c|}{\multirow{1}{*}{\textbf{\checkmark }}}
&\multicolumn{1}{|c|}{\multirow{1}{*}{\textbf{\checkmark }}} \\
% \cline{2-6}
% 
% \multicolumn{1}{|c|}{\multirow{1}{*}{\textbf{FTF}}}
% 
% &\multicolumn{1}{|c|}{\multirow{1}{*}{\textbf{2}}}
% 
% &\multicolumn{1}{|c|}{\multirow{1}{*}{\textbf{\checkmark}}}
% % 
% &\multicolumn{1}{|c|}{\multirow{1}{*}{\textbf{\checkmark}}}
% % 
% &\multicolumn{1}{|c|}{\multirow{1}{*}{\textbf{\checkmark }}}
% % 
% &\multicolumn{1}{|c|}{\multirow{1}{*}{\textbf{\checkmark }}} \\
\hline

\end{tabular}
\label{tab:schedulers}
\end{table}

\section{Case Studies}
\label{sec:case_studies}
We begin by analyzing the performance of the models listed in~\autoref{tab:unit_models}. 
To collect performance data, we use ONNX Runtime Generative AI (OGA) for LLMs and ONNX Runtime (ORT) for all other models.
The resulting profiling data populates the simulator's latency database, which we use to simulate the workload scenarios from~\autoref{tab:scenarios} and evaluate the impact of different scheduling strategies on \rtgen workloads running on emerging heterogeneous platforms.

\subsection{Workload Characterization}
\label{subsec:workload_char}
% Discuss the collected profiling results of the selected work: Discuss the preferred backend for every model, Ablation Study showing the performance gains of leveraging heterogeneity for the LLMs.
We characterize the performance of the models in~\autoref{tab:unit_models} and present the results in ~\autoref{fig:rt-stx} and ~\autoref{fig:llm-profiling}. 
As shown in~\autoref{fig:rt-stx} and~\autoref{fig:llm-profiling}, no single backend consistently offers the best performance across all models.
% 
% We observe in ~\autoref{fig:rt-stx} and ~\autoref{fig:llm-profiling} that there is not a unique backend that yields the best performance for all the models. 
% %  

\betterparagraph{CNN-based Models} The super-resolution (SR) network incurs the highest latency when executed on the CPU, requiring 3.78 ms on average.  
Running the SR on the GPU reduces the inference latency by 14.6\%.
% Running the SR on the GPU slightly improves the inference latency with a 1.17$\times$ speedup.
% 
Nevertheless, deploying the SR network on the NPU achieves a significant speedup compared to the GPU and CPU. The NPU reduces the latency of the SR by 81.9\% and 84.5\% with respect to the GPU and CPU, and runs the SR inference in 0.59 ms on average. 

A similar performance trend is observed for the object detection (OD) model.
After NPU acceleration, the latency of the OD inference is 5.23 ms on average, a 42.9\% and a 62.7\% decrease compared to the GPU and CPU latencies. 
% 
% 0.59 ms, reducing the GPU and CPU inference latency by 81.9\% and 84.5\% respectively. 
%  
% 
In contrast, the segmentation (Seg) model achieves its lowest latency on the GPU which is 10.2\% faster than the NPU.   
% The inference latency of the Seg model on the GPU is 12.35 ms which is 10.2\% lower than the NPU latency. 

% we do not observe a similar trend for the segmentation (Seg) network. 
% %  
% Unlike SR, Seg runs most efficiently on the GPU with a slight 1.11$\times$ speedup over NPU. 
%  

\betterparagraph{Transformer-based Models} In~\autoref{fig:llm-profiling}, we observe that the encoder's inference latency is the lowest for sequence lengths of 16 and 32 when the model is deployed on the NPU. The NPU latency is 31.9\% and 10.9\% lower than the GPU latency for an input sequence of length 16 and 32 respectively. 
As the sequence length increases, we notice that the NPU latency of the encoder becomes greater than the encoder's GPU latency. 
At a sequence of 1024, the GPU inference latency is 22.3\% lower than the NPU latency. 
These results indicate that the encoder's optimal backend selection is input-dependent and varies with sequence length.

As for the LLM, we observe that the backend resulting in the lowest inference latency varies across the generation stages (perfill and decode). 
% We observe a heterogeneity in the lowest latency backend for LLMs, a similar trend to the one observed for the real-time models. 
% %  
For the prefill stage, we notice in~\autoref{fig:llm-profiling} that running inference on the NPU results in an average speedup of 3.0$\times$ across all input sequence lengths compared to running on the GPU.  
On the other hand, running the decode stage on the GPU results in a 7.5$\times$ speedup over the NPU on average across varying context lengths. 
Since the prefill phase is a compute intensive stage and the decode phase is a memory and bandwidth (BW) intensive stage, the collected profiling data indicate the NPU is better suited for compute-bound workloads while the GPU is more suited for memory/BW-bound workloads on our evaluated system. 

The profiling results reveal heterogeneity in backend preference both across different models and between stages of the same model.

This emphasizes the importance and advantages of heterogeneous platforms that provide multiple backend each tailored to a specific workload. Moreover, it indicates that heterogeneous systems are a great fit to run \rtgen workloads characterized by heterogeneous models. 

\subsection{\rtgen Scheduling}
\label{subsec:scheduling_impact}

% \insertWideFigure{deadline-ttft-tpt}{\TODO{Frame Drop, and Frame Drop Rete, Heterogeneous Backend, Parallel Execution, Deadline Awareness.}The impact of various scheduling policies on performance. FCFS-Static and EDF-Static do not leverage heterogeneity; they only schedule a single layer at a time on its preferred backend. The empty bars in the in the top row correspond to a 0\% violation rate. The empty bars in the middle and bottom rows indicate workload starvation and/or incomplete simulation.}

As discussed in~\autoref{subsec:workload_char}, \rtgen models are heterogeneous and have diverse backend preferences. 
In addition, \rtgen workloads can exhibit dynamic behaviors due to dynamic input shapes, have strict deadlines, and run multiple models concurrently. 

To evaluate how \rtgen workload characteristics interact with scheduling decisions on heterogeneous systems, we simulate four scenarios with five different scheduling policies.
For every scenario, we analyze the models' deadline violation rate, the LLM's time-to-first-token (TTFT), and its time-per-token (TPT). We present the results in~\autoref{tab:results}.
We also evaluate the effects of varying the input sequence length in~\autoref{fig:inp-seq-len-perf}. 

\subsubsection{Scheduling Algorithm}
For scenario A, we observe identical performance under all scheduling policies. This is expected because scenario A (AI chatbot assistant) does not have any real-time constraints. It also does not exhibit any model concurrency which simplifies the scheduling decision, because the RAG pipeline does not have to compete for resources with other models. Hence, the scheduling policy does not impact the performance of scenario A. 
On the other hand, the performance of the remaining scenarios is significantly affected by the scheduling policy. 

\betterparagraph{Earliest Deadline} We notice that using an earliest deadline first (EDF) policy will always starve the generative LLM in all the workloads. 
EDF consistently prioritizes the model with the closest deadline, preempting LLM execution in favor of real-time tasks.
For example, in scenario C, the EDF-AOT scheduler stops the LLM's prefill execution to schedule the SR-120 on the NPU.
This behavior leads to repeated preemption of the LLM, resulting in starvation and failure to complete execution.
% as we show in~\autoref{fig:challenges}.
%  
The case is similar for the EDF-DYN scheduler. Despite scheduling the models' layers dynamically on the HW, it still prioritizes the model with the earliest deadline and schedules it on the best HW backend available. Since the latency of a single LLM layer in scenario C (98.62 ms) is larger than the largest frame duration (16.67 ms), the LLM will be stopped in favor of a model with a closer deadline, leading to the starvation of the LLM task. 
This indicates that \textit{only considering deadline violation requirements greatly degrades the LLM's performance.} 

\betterparagraph{First Come First Serve} While servicing inference requests in arrival order improves the LLM’s TTFT and TPT relative to EDF strategies, it comes at the cost of increased deadline violations.
We observe that applying a FCFS scheduler leads to a TTFT performance that matches the standalone case. This observation is consistent across all scenarios B, C, and D.
Nevertheless, a FCFS scheduling policy results in an increased deadline violation rate.
For scenarios C and D, the deadline violation rates is 57.65\% on average when applying a FCFS strategy.
Under the FCFS-AOT policy in scenario C, the LLM achieves TTFT and TPT performance comparable to the (scenario A) but with a 62.6\% deadline violation rate. 
Switching to FCFS-DYN that dynamically schedule layers on the best available HW decreases the deadline violation rate to 39.9\%. 
As shown in~\autoref{tab:model_drop_rate}, the deadline violation rate of SR-60 and Seg models significantly decreased to 0.5\% and 0.0\% respectively when applying dynamic scheduling.  
Nevertheless, the deadline violation rate of SR-120 increases to 77.4\%, as well as the LLM's TPT increases by 25.5\% in scenario C under FCFS-DYN.
For scenario D, we observe that applying FCFS-DYN does not reduce the deadline rate which remains at 66.5\%, but increases the TPT by 86.9\%.
We observe in~\autoref{tab:model_drop_rate} that applying dynamic HW selection on scenario D shifted the deadline violations from the Seg model to the OD model. 
Therefore, utilizing FCFS to schedule \rtgen workloads leads to high deadline violation rates (57.6\% on average). While dynamic scheduling might limit the deadline violation problem, it is scenario and model dependent and can result in an increase in TPT.

\betterparagraph{GenAI Aware} Despite showing promising TTFT and TPT, FCFS schedulers fail to account for the real-time requirements of \rtgen workloads. 
Similarly, EDF policies cannot accommodate the long latencies of the generative LLMs. 
Therefore, \rtgen workloads require dynamic schedulers that are both real-time and generative AI aware. 
We observe that our simple heterogeneous policy FTF does not starve the LLM, achieves a TTFT that matches the standalone performance, and  has a lower average deadline violation rate than FCFS. 
For scenarios B, C, and D, the TTFT is respectively 127.4 ms, 1577.9 ms, and 1577.9 ms matching the standalone scenario when running under FTF.

By assigning high priority to the first token generation, the FTF scheduler leverages the heterogeneous SoC to avoid EDF's repeated preemption of the LLM and schedule the inference while still meeting real-time constraints for other models. 
This is possible because FTF explicitly treats the LLM’s TTFT as a high-priority deadline, preventing preemption during the prefill stage. 
Such prioritization is necessary due to the large gap between LLM layer latencies and the frame durations of real-time models, e.g., 98.6 ms vs. 16.7 ms in scenario C, which makes the LLM particularly vulnerable to starvation under standard EDF. 
In contrast, during the decode stage, the LLM’s per-layer latency becomes more comparable to that of other models, allowing for more balanced scheduling without excessive preempting under a conventional EDF.

% By assigning high priority to the first token generation, the FTF scheduler leverages the heterogeneous SoC to avoid EDF's repeated preemption of the LLM and schedule the inference while still meeting real-time constraints for other models.
% 
% By assigning a high priority to generating the first token, the FTF scheduler leverages the heterogeneous SoC to running the LLM inference while still trying to meet the deadlines of the other models. 
%  
Under FTF, the deadline violation rates of scenarios C and D decreases by 47.8\% on average compared to FCFS.
Nevertheless, the TPT significantly increases under TFT compared to the standalone case.
The TPT of scenario C increases by $5.7\times$, and that of scenario D by $7.4\times$.
The increase in TPT results in an increase in the decode stage and end-to-end latencies of the LLM inference especially as the number of generated tokens increases. 
These results further emphasize the importance of a workload-aware scheduler that can dynamically adapt to balance real-time constraints, multi-modality, and the long, input-dependent latencies of generative models.
% Hence, we motivate the importance of having a workload-aware scheduler to dynamically balance the real-time constraints, the multi-modality execution, and the generative model's fluctuating latencies. 
% 

\subsubsection{Impact of the Sequence Length} 
As discussed in~\autoref{subsec:rt-models}, \rtgen workloads can involve dynamic tensor shapes. 
Varying input prompt lengths in LLMs represent a form of tensor shape dynamicity common in generative workloads.
% The varying input prompt or sequence length of the LLM is an example of tensor shape dynamicity found in generative workloads. 
%  
Therefore we study how tensor shape dynamicity affect the scheduling of \rtgen workloads.
We simulate scenario D with varying sequence lengths and list the results in~\autoref{fig:inp-seq-len-perf}. 
The results in~\autoref{fig:inp-seq-len-perf} (a) indicate an increase in the deadline violation rate under the FTF scheduler from 1.7\% for an input sequence of 32 tokens to 10.7\% for an input of 2048 tokens.
As we increase the input sequence length, the prefill latency increases as~\autoref{fig:llm-profiling} shows.
Consequently, the LLM is occupying the available resources, the NPU in this case, for a longer time which prevents other models from being scheduled and leads to deadline violations. 
% 

% In addition, we observe an increase in TPT compared to the standalone TPT for all the sequence lengths. This increase is caused by running some decode layers on the NPU (the non-preferred backend which is $7.5\times$ slower than the GPU in decode) as can be seen in~\autoref{fig:backend_partition_tpt_scenario_D}. This increase is also a result of additional waiting in the inference queues. 
% %  
% More interestingly, we notice that TPT for a 32 and 64 token input is in the same range if the TPT of a 2048 token input.
% %  
% This is typically not expected because the TPT usually scales with the context length. 
% %  
% Nevertheless, the high TPT for the 32 and 64 token cases results from running 95.1\% of the decode layers on the NPU as we show in~\autoref{fig:backend_partition_tpt_scenario_D}. For the other sequence lengths, only 37.4\% of layers run on the NPU on average. This is beacause of the varying latencies of the input-dependent prefill stage. 
% % 
% Based on these observations, we further motivate the need for a workload aware scheduler that takes into account the dynamic latencies of models caused by dynamic input shapes. 
% 
In addition, we observe an increase in TPT across all sequence lengths compared to the standalone case. This increase is caused by two primary factors. 
First, a portion of the decode layers are executed on the NPU, which is not the preferred backend for this stage. 
As shown in~\autoref{fig:llm-profiling}, the NPU is $7.5\times$ slower than the GPU for decode operations, leading to a substantial increase in total inference time. Second, the increase is further exacerbated by additional waiting times in the inference queues due to backend contention.

More interestingly, we find that the TPT for 32- and 64-token inputs falls within the same range as the TPT for 2048-token inputs. 
This observation is unexpected, as TPT typically increases with context length. 
However, in this case, the high TPT for short sequences stems from aggressive offloading of decode layers to the NPU. Specifically, 95.1\% of decode layers for the 32- and 64-token inputs are dispatched to the NPU, as shown in~\autoref{fig:backend_partition_tpt_scenario_D}. In contrast, for longer sequence lengths, only 37.4\% of decode layers are executed on the NPU on average. This difference is attributed to the latency variability introduced by the input-dependent prefill stage, which affects backend availability at scheduling time.
Based on these observations, we further motivate the need for a workload-aware dynamic scheduler that accounts for the dynamic execution behavior of generative models. 
% In particular, such a scheduler must consider how input-dependent latency fluctuations influence backend assignment and overall inference performance.

\insertFigure{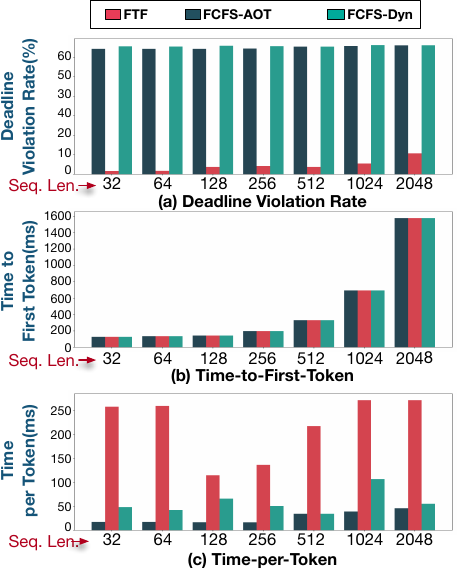}{The impact of the input sequence length on the deadline violation rate, time-to-first-token, and time-per-token of scenario D.}
\insertFigure{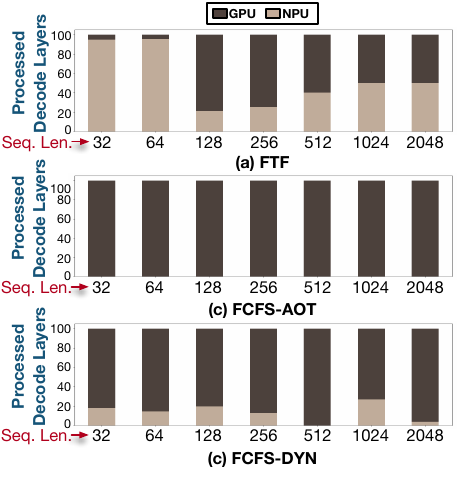}{The percentage of LLM layers processed on the GPU and NPU during 32 decode iterations for the video conference scenario (scneario D in~\autoref{tab:scenarios}.}

\betterparagraph{Summary} Our case studies highlight the performance heterogeneity of \rtgen workloads and the need for intelligent scheduling on heterogeneous systems.
We show that preferred backends vary across models and even within different stages of a single model, as is the case with generative LLMs. 
This behavior is further complicated by dynamic input shapes, which shift backend suitability during runtime. 
We find that conventional scheduling policies such as EDF and FCFS fail to balance the conflicting demands of low-latency token generation and real-time deadline compliance.
While FCFS improves LLM TTFT and TPT, it does so at the cost of high deadline violation rates. 
EDF policies, on the other hand, prioritize real-time models but starve the LLM entirely. 
Our GenAI-aware FTF scheduler strikes a better balance, achieving standalone-like TTFT with reduced deadline violations, but introduces higher TPT due to backend contention and decode-stage bottlenecks. 
We also demonstrate that input sequence length significantly impacts scheduling behavior, revealing that shorter sequences can suffer disproportionately due to resource contention and backend assignment. 
These findings underscore the need for a workload-aware scheduler that dynamically adapts to input characteristics, model stage, and system heterogeneity to meet the diverse demands of \rtgen workloads.

\begin{table}
\caption{The impact of various scheduling policies from ~\autoref{tab:schedulers} on the deadline violation rate, time-to-first-token (TTFT), and time per token (TPT) of the workload scenarios defined in~\autoref{tab:scenarios}. }
\scriptsize 
\center
\begin{tabular}{|l|l|l|l|l|l|l|l|l|l|}

\hline
\multicolumn{1}{|c|}{\multirow{3}{*}{\textbf{Scenario}}}
&\multicolumn{1}{|c|}{\multirow{3}{*}{\textbf{Scheduler}}}
&\multicolumn{3}{|c|}{\multirow{1}{*}{\textbf{Performance Metrics}}}\\
\cline{3-5}
% 
% &\multicolumn{1}{|c|}{\multirow{1}{*}{\textbf{Backends}}}
% 
&
&\multicolumn{1}{|c|}{\multirow{1}{*}{\textbf{Deadline}}}
&\multicolumn{1}{|c|}{\multirow{1}{*}{\textbf{Time to }}} 
&\multicolumn{1}{|c|}{\multirow{1}{*}{\textbf{Time per}}}\\

& 
&\multicolumn{1}{|c|}{\multirow{1}{*}{\textbf{Viol. Rate(\%)}}}
&\multicolumn{1}{|c|}{\multirow{1}{*}{\textbf{First Token(ms)}}} 
&\multicolumn{1}{|c|}{\multirow{1}{*}{\textbf{Token(ms)}}}\\
\hline 
% Senario A 
\multicolumn{1}{|c|}{\multirow{5}{*}{\textbf{Scenario A}}} 
&\multicolumn{1}{|c|}{\multirow{1}{*}{\textbf{EDF-AOT}}}
&\multicolumn{1}{|c|}{\multirow{1}{*}{-}}
&\multicolumn{1}{|c|}{\multirow{1}{*}{\textbf{1577.9}}}
&\multicolumn{1}{|c|}{\multirow{1}{*}{\textbf{45.9}}}\\
\cline{2-5}

&\multicolumn{1}{|c|}{\multirow{1}{*}{\textbf{EDF-Dyn}}}
&\multicolumn{1}{|c|}{\multirow{1}{*}{-}}
&\multicolumn{1}{|c|}{\multirow{1}{*}{\textbf{1577.9}}}
&\multicolumn{1}{|c|}{\multirow{1}{*}{\textbf{45.9}}}\\
\cline{2-5}

&\multicolumn{1}{|c|}{\multirow{1}{*}{\textbf{FCFS-AOT}}}
&\multicolumn{1}{|c|}{\multirow{1}{*}{-}}
&\multicolumn{1}{|c|}{\multirow{1}{*}{\textbf{1577.9}}}
&\multicolumn{1}{|c|}{\multirow{1}{*}{\textbf{45.9}}}\\
\cline{2-5}

&\multicolumn{1}{|c|}{\multirow{1}{*}{\textbf{FCFS-Dyn}}}
&\multicolumn{1}{|c|}{\multirow{1}{*}{-}}
&\multicolumn{1}{|c|}{\multirow{1}{*}{\textbf{1577.9}}}
&\multicolumn{1}{|c|}{\multirow{1}{*}{\textbf{45.9}}}\\
\cline{2-5}

&\multicolumn{1}{|c|}{\multirow{1}{*}{\textbf{FTF}}}
&\multicolumn{1}{|c|}{\multirow{1}{*}{-}}
&\multicolumn{1}{|c|}{\multirow{1}{*}{\textbf{1577.9}}}
&\multicolumn{1}{|c|}{\multirow{1}{*}{\textbf{45.9}}}\\
\hline 

\multicolumn{1}{|c|}{\multirow{5}{*}{\textbf{Scenario B}}} 
&\multicolumn{1}{|c|}{\multirow{1}{*}{\textbf{EDF-AOT}}}
&\multicolumn{1}{|c|}{\multirow{1}{*}{\textbf{0.0}}}
&\multicolumn{1}{|c|}{\multirow{1}{*}{\textbf{-}}}
&\multicolumn{1}{|c|}{\multirow{1}{*}{\textbf{-}}}\\
\cline{2-5}

&\multicolumn{1}{|c|}{\multirow{1}{*}{\textbf{EDF-Dyn}}}
&\multicolumn{1}{|c|}{\multirow{1}{*}{\textbf{0.0}}}
&\multicolumn{1}{|c|}{\multirow{1}{*}{\textbf{-}}}
&\multicolumn{1}{|c|}{\multirow{1}{*}{\textbf{-}}}\\
\cline{2-5}

&\multicolumn{1}{|c|}{\multirow{1}{*}{\textbf{FCFS-AOT}}}
&\multicolumn{1}{|c|}{\multirow{1}{*}{\textbf{0.3}}}
&\multicolumn{1}{|c|}{\multirow{1}{*}{\textbf{127.4}}}
&\multicolumn{1}{|c|}{\multirow{1}{*}{\textbf{39.0}}}\\
\cline{2-5}

&\multicolumn{1}{|c|}{\multirow{1}{*}{\textbf{FCFS-Dyn}}}
&\multicolumn{1}{|c|}{\multirow{1}{*}{\textbf{0.0}}}
&\multicolumn{1}{|c|}{\multirow{1}{*}{\textbf{127.4}}}
&\multicolumn{1}{|c|}{\multirow{1}{*}{\textbf{39.0}}}\\
\cline{2-5}

&\multicolumn{1}{|c|}{\multirow{1}{*}{\textbf{FTF}}}
&\multicolumn{1}{|c|}{\multirow{1}{*}{\textbf{0.0}}}
&\multicolumn{1}{|c|}{\multirow{1}{*}{\textbf{127.4}}}
&\multicolumn{1}{|c|}{\multirow{1}{*}{\textbf{39.0}}}\\
\hline 

\multicolumn{1}{|c|}{\multirow{5}{*}{\textbf{Scenario C}}} 
&\multicolumn{1}{|c|}{\multirow{1}{*}{\textbf{EDF-AOT}}}
&\multicolumn{1}{|c|}{\multirow{1}{*}{\textbf{0.0}}}
&\multicolumn{1}{|c|}{\multirow{1}{*}{\textbf{-}}}
&\multicolumn{1}{|c|}{\multirow{1}{*}{\textbf{-}}}\\
\cline{2-5}

&\multicolumn{1}{|c|}{\multirow{1}{*}{\textbf{EDF-Dyn}}}
&\multicolumn{1}{|c|}{\multirow{1}{*}{\textbf{0.0}}}
&\multicolumn{1}{|c|}{\multirow{1}{*}{\textbf{-}}}
&\multicolumn{1}{|c|}{\multirow{1}{*}{\textbf{-}}}\\
\cline{2-5}

&\multicolumn{1}{|c|}{\multirow{1}{*}{\textbf{FCFS-AOT}}}
&\multicolumn{1}{|c|}{\multirow{1}{*}{\textbf{62.6}}}
&\multicolumn{1}{|c|}{\multirow{1}{*}{\textbf{1577.9}}}
&\multicolumn{1}{|c|}{\multirow{1}{*}{\textbf{45.9}}}\\
\cline{2-5}

&\multicolumn{1}{|c|}{\multirow{1}{*}{\textbf{FCFS-Dyn}}}
&\multicolumn{1}{|c|}{\multirow{1}{*}{\textbf{39.9}}}
&\multicolumn{1}{|c|}{\multirow{1}{*}{\textbf{1577.9}}}
&\multicolumn{1}{|c|}{\multirow{1}{*}{\textbf{61.6}}}\\
\cline{2-5}

&\multicolumn{1}{|c|}{\multirow{1}{*}{\textbf{FTF}}}
&\multicolumn{1}{|c|}{\multirow{1}{*}{\textbf{44.4}}}
&\multicolumn{1}{|c|}{\multirow{1}{*}{\textbf{1577.9}}}
&\multicolumn{1}{|c|}{\multirow{1}{*}{\textbf{263.1}}}\\
\hline 

\multicolumn{1}{|c|}{\multirow{5}{*}{\textbf{Scenario D}}} 
&\multicolumn{1}{|c|}{\multirow{1}{*}{\textbf{EDF-AOT}}}
&\multicolumn{1}{|c|}{\multirow{1}{*}{\textbf{0.0}}}
&\multicolumn{1}{|c|}{\multirow{1}{*}{\textbf{-}}}
&\multicolumn{1}{|c|}{\multirow{1}{*}{\textbf{-}}}\\
\cline{2-5}

&\multicolumn{1}{|c|}{\multirow{1}{*}{\textbf{EDF-Dyn}}}
&\multicolumn{1}{|c|}{\multirow{1}{*}{\textbf{0.0}}}
&\multicolumn{1}{|c|}{\multirow{1}{*}{\textbf{-}}}
&\multicolumn{1}{|c|}{\multirow{1}{*}{\textbf{-}}}\\
\cline{2-5}

&\multicolumn{1}{|c|}{\multirow{1}{*}{\textbf{FCFS-AOT}}}
&\multicolumn{1}{|c|}{\multirow{1}{*}{\textbf{66.5}}}
&\multicolumn{1}{|c|}{\multirow{1}{*}{\textbf{1577.9}}}
&\multicolumn{1}{|c|}{\multirow{1}{*}{\textbf{84.0}}}\\
\cline{2-5}

&\multicolumn{1}{|c|}{\multirow{1}{*}{\textbf{FCFS-Dyn}}}
&\multicolumn{1}{|c|}{\multirow{1}{*}{\textbf{66.5}}}
&\multicolumn{1}{|c|}{\multirow{1}{*}{\textbf{1577.9}}}
&\multicolumn{1}{|c|}{\multirow{1}{*}{\textbf{85.8}}}\\
\cline{2-5}

&\multicolumn{1}{|c|}{\multirow{1}{*}{\textbf{FTF}}}
&\multicolumn{1}{|c|}{\multirow{1}{*}{\textbf{20.8}}}
&\multicolumn{1}{|c|}{\multirow{1}{*}{\textbf{1577.9}}}
&\multicolumn{1}{|c|}{\multirow{1}{*}{\textbf{339.9}}}\\
\hline 

\end{tabular}
\label{tab:results}
\end{table}
% \insertFigure{inp-seq-len-perf}{The impact of the input sequence length on the deadline violation rate, time-to-first-token, and time-per-token of scenario D.}
% \insertFigure{backend_partition_tpt_scenario_D}{The percentage of LLM layers processed on the GPU and NPU during 32 decode iterations for the video conference scenario (scneario D in~\autoref{tab:scenarios}.}

\begin{table}
\caption{The deadline violation rate per model for scenarios C and D.}
\scriptsize 
\center
\begin{tabular}{|l|l|l|l|l|l|l|l|l|l|}

\hline

\multicolumn{1}{|c|}{\multirow{2}{*}{\textbf{Scenario}}}
& \multicolumn{1}{|c|}{\multirow{2}{*}{\textbf{Models}}}
& \multicolumn{3}{|c|}{\multirow{1}{*}{\textbf{Schedulers}}}\\
\cline{3-5}
&
&\multicolumn{1}{|c|}{\multirow{1}{*}{\textbf{FCFS-AOT}}}
&\multicolumn{1}{|c|}{\multirow{1}{*}{\textbf{FCFS-DYN}}}
&\multicolumn{1}{|c|}{\multirow{1}{*}{\textbf{FTF}}}\\
\hline
\multicolumn{1}{|c|}{\multirow{3}{*}{\textbf{Gaming II}}}
&\multicolumn{1}{|c|}{\multirow{1}{*}{\textbf{SR-120}}}
&\multicolumn{1}{|c|}{\multirow{1}{*}{\textbf{53.1\%}}}
&\multicolumn{1}{|c|}{\multirow{1}{*}{\textbf{77.4\%}}}
&\multicolumn{1}{|c|}{\multirow{1}{*}{\textbf{8.3\%}}}\\
\cline{2-5}
&\multicolumn{1}{|c|}{\multirow{1}{*}{\textbf{SR-60}}}
&\multicolumn{1}{|c|}{\multirow{1}{*}{\textbf{99.5\%}}}
&\multicolumn{1}{|c|}{\multirow{1}{*}{\textbf{0.5\%}}}
&\multicolumn{1}{|c|}{\multirow{1}{*}{\textbf{95.8\%}}}\\
\cline{2-5}
&\multicolumn{1}{|c|}{\multirow{1}{*}{\textbf{Seg}}}
&\multicolumn{1}{|c|}{\multirow{1}{*}{\textbf{46.5\%}}}
&\multicolumn{1}{|c|}{\multirow{1}{*}{\textbf{0.0\%}}}
&\multicolumn{1}{|c|}{\multirow{1}{*}{\textbf{69.8\%}}}\\
\hline 

\multicolumn{1}{|c|}{\multirow{3}{*}{\textbf{Video Conference}}}
&\multicolumn{1}{|c|}{\multirow{1}{*}{\textbf{SR}}}
&\multicolumn{1}{|c|}{\multirow{1}{*}{\textbf{99.4\%}}}
&\multicolumn{1}{|c|}{\multirow{1}{*}{\textbf{99.5\%}}}
&\multicolumn{1}{|c|}{\multirow{1}{*}{\textbf{16.1\%}}}\\
\cline{2-5}
&\multicolumn{1}{|c|}{\multirow{1}{*}{\textbf{Seg}}}
&\multicolumn{1}{|c|}{\multirow{1}{*}{\textbf{46.5\%}}}
&\multicolumn{1}{|c|}{\multirow{1}{*}{\textbf{0.0\%}}}
&\multicolumn{1}{|c|}{\multirow{1}{*}{\textbf{0.0}}}\\
\cline{2-5}
&\multicolumn{1}{|c|}{\multirow{1}{*}{\textbf{OD}}}
&\multicolumn{1}{|c|}{\multirow{1}{*}{\textbf{53.9\%}}}
&\multicolumn{1}{|c|}{\multirow{1}{*}{\textbf{99.5\%}}}
&\multicolumn{1}{|c|}{\multirow{1}{*}{\textbf{16.1\%}}}\\
\hline

\end{tabular}
\label{tab:model_drop_rate}
\end{table}

\section{Related Works}
\label{sec:realted_works}

\betterparagraph{LLM Scheduling}Many existing scheduling works \cite{sheng2023flexgen, kamahori2024fiddler,lee2024infinigen, heo2024neupims} optimize the deployment of generative LLMs. 
Flexgen \cite{sheng2023flexgen} proposes a throughput oriented schedule to enable running LLMs on a single GPU with limited memory.  
Flexgen statically partitions the layers of LLMs across CPU and GPU memory, and schedules the GPU computations to maximize weight reuse.
Nevertheless, Flexgen does not account for the multi-model inference scenarios, which makes its static scheduling approach unsuitable for emerging workloads, described in ~\autoref{sec:genai_rt}, that concurrently running LLMs with real-time models. 
NeuPIMS~\cite{heo2024neupims} proposes a heterogeneous architecture that integrates NPUs with processing in memory cores (PIMs) to accelerate batched generative LLM inference.
NeuPIMS scheduler partitions the bathes into sub-batches to parallelize the execution of the multi-head attention layer across the heterogeneous backend. 
Nonetheless, NeuPIMS only focuses on LLM serving and does not investigate the LLM inference scheduling problem in a real-time multi-model settings. 
Unlike previous works~\cite{sheng2023flexgen, kamahori2024fiddler,lee2024infinigen, heo2024neupims}, our work studies the performance of generative LLM inference on heterogeneous systems in real-time application scenarios. 
We evaluate the importance of the scheduling policy in managing the challenges of real-time generative multi-model workloads. 

\betterparagraph{Scheduling on Heterogeneous Systems}Previous works~\cite{kim2023dream,ghodrati2020planaria, kwon2021herald, bouzidi2023map} investigate the scheduling problem on heterogeneous systems. However, their characterization and approach focus on a limited set of workloads ~\cite{bouzidi2023map} or a specific class of backend~\cite{kim2023dream}. 
DREAM~\cite{kim2023dream} is a state-of-the-art dynamic scheduler tailored for real-time multi-model (RTMM) workloads. 
DREAM quantifies the requirements of RTMM workloads into metrics which are used to drive the scheduling decision. It employs tunable parameters to effectively adapt to the workload's dynamicity. 
In addition, DREAM focus on RTMM workloads based on AR/VR applications running on heterogeneous dataflow accelerator backend.
Our work extends on these efforts by analyzing the impact of scheduling on a new class of emerging models integrating RTMM with generative models deployed on NPU-GPU systems. 
Map and Conquer~\cite{bouzidi2023map} offers an optimization framework to statically partition and schedule single model neural network inference on GPU-NPU systems.
Because the proposed approach is tailored to single model inference, it does not account for the dynamic change in resource availability that results from multi-model workloads. 
Our work motivates the need for a dynamic scheduler when running real-time generative workloads.
\section{Conclusion}
\label{sec:conclusion}
The emergence of \rtgen workloads poses significant challenges for on-device execution.
While heterogeneous SoCs' characteristics seem very promising for such emerging workloads, their scheduling space is exponentially large. 
In this work, we explore the impact of the scheduling on the performance horizon of emerging \rtgen workloads. 
We define four realistic scenarios and thoroughly characterize their performance under five scheduling policies. 
Our study shows that the scheduling policy significantly affects the deadline violation rate, TTFT, and TPT of the \rtgen scenarios, and it motivates the need for workload- aware scheduling strategies that can dynamically adapt to backend availability, model heterogeneity, and the characteristics of \rtgen workloads.

\begin{comment}
The adoption of tightly integrated CPU-NPU-GPU SoCs in recent AI PCs enabled emerging workloads integrating GenAI and real-time workloads in the same application. 
%  
This class of emerging workloads present multiple challenges such as model heterogeneity, multi-model concurrency, real-time constraints, and dynamic tensor shapes. 
% 
To better understand the performance of real-time and generative multi-model workloads, we define four realistic scenarios inspired by industry applications. 
% 
We characterize the real-time and generative workloads used to define the scenarios. Our experiments show heterogeneity in the preferred backend of each workload.
% 
We also show the impact the scheduling algorithm can have on the performance of the defined scenarios.
%  
Our analysis confirms the importance of HW heterogeneity to obtain optimal performance. Moreover, it implies that we need dynamic schedulers that are tailored to the characteristics of the emerging real-time generative workloads.
\end{comment}
%\section*{Acknowledgments}
%HK: Shouldn't be included in the submission for review

%%%%%%%%% -- BIB STYLE AND FILE -- %%%%%%%%
%\bibliographystyle{IEEEtranS}
\bibliographystyle{plain}
\bibliography{ref}
%%%%%%%%%%%%%%%%%%%%%%%%%%%%%%%%%%%%
\end{document}